\documentclass[journal]{IEEEtran}

\usepackage{amsmath,amsfonts,amssymb}
\usepackage{mathtools}
\usepackage{array}
\usepackage{booktabs}
\usepackage{multirow}
\usepackage{graphicx}
\usepackage{xcolor}
\usepackage{cite}
\usepackage{url}
\usepackage{textcomp}
\usepackage{stfloats}
\usepackage{microtype}

\begin{document}

\title{Towards Compact Unified Multimodal Tracking: Synergizing Knowledge Distillation with Structural Pruning}

\author{Yuqi~Li,
        Yuedong~Tan,
        Huiran~Duan,
        Weilun~Feng,
        Chuanguang~Yang,
        Zhulin~An,
        Zongwei~Wu,\\
        Shiping~Wen,~\IEEEmembership{Senior~Member,~IEEE},
        Tingwen~Huang,~\IEEEmembership{Fellow,~IEEE},
        and~Yingli~Tian*,~\IEEEmembership{Fellow,~IEEE}%
\thanks{Yuqi Li and Yuedong Tan contributed equally to this work.}%
\thanks{Corresponding author: Yingli~Tian (e-mail: ytian@ccny.cuny.edu).}%
\thanks{Yuqi Li, Huiran Duan, and Yingli Tian are with The City College of New York, New York, NY, USA.}%
\thanks{Yuedong Tan is with INSAIT, Sofia University “St. Kliment Ohridski”, Sofia, Bulgaria, and Julius-Maximilians-Universität Würzburg, Würzburg, Germany.}%
\thanks{Weilun Feng, Chuanguang Yang, and Zhulin An are with the State Key Laboratory of AI Safety, Institute of Computing Technology, Chinese Academy of Sciences, Beijing, China. Weilun Feng is also with the University of Chinese Academy of Sciences.}%
\thanks{Zongwei Wu is with the Julius-Maximilians-Universität Würzburg, Würzburg, Germany.}%
\thanks{Shiping Wen and Tingwen Huang are with Shenzhen University of Advanced Technology, Shenzhen, China.}%
}

\maketitle

\begin{abstract}
Unified multimodal object tracking has achieved remarkable robustness by leveraging complementary sensor data (e.g., RGB, Thermal, Depth), yet the heavy computational burden of state-of-the-art models hinders their deployment on resource-constrained edge devices. In this work, we identify the prediction head as a critical but often overlooked efficiency bottleneck. By strategically streamlining the decoder architecture, we unlock the potential for real-time inference but simultaneously introduce a capacity gap between the lightweight student and the heavy teacher. To resolve this, we conduct a systematic analysis of 17 distillation strategies and introduce a Dual-Alignment Distillation framework. Our key insight is that effective compression requires decoupling knowledge transfer into two complementary streams: (1) \textit{Spatial Representation Alignment}, which employs feature distillation to sharpen the student's spatial focus on foreground targets ("Where to track"); and (2) \textit{Semantic Distribution Alignment}, which utilizes logit-based distillation to align decision boundaries and transfer discriminative dark knowledge ("What to track"). Extensive experiments across five benchmarks demonstrate that our approach significantly outperforms complex state-of-the-art methods. Notably, our distilled model achieves 91.5\% MPR on RGBT234 and operates at 54 FPS on a single RTX 4090, representing a 5$\times$ speedup over the teacher model while maintaining superior accuracy.
\end{abstract}


\section{Introduction}

\IEEEPARstart{V}{isual} object tracking has made substantial advances in recent years, driven by the development of sophisticated spatiotemporal models ~\cite{ostrack,hhtrack,mixformer}. However, despite the availability of large-scale datasets, trackers relying solely on the RGB modality remain susceptible to environmental challenges such as poor illumination, extreme occlusion, and motion blur. To mitigate these inherent limitations, the field has increasingly pivoted toward Multimodal Object Tracking, fusing RGB data with complementary information from thermal, depth, or event sensors to achieve robust performance in complex scenarios~\cite{brodermann2025cafuser}.

\begin{figure}[!t]
\centering
\includegraphics[width=\linewidth,keepaspectratio]{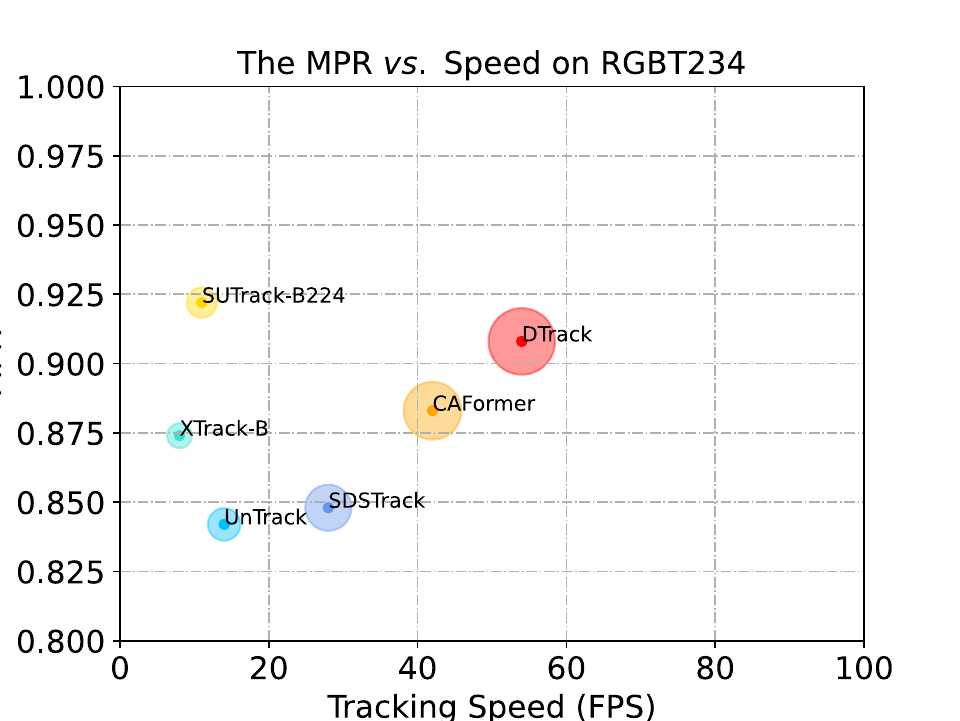}
\caption{\textbf{Speed \& Performance comparison with state-of-the-art methods on RGBT234 dataset:} Our method (red bubble) achieves an optimal trade-off between accuracy and efficiency, attaining state-of-the-art precision comparable to the heavyweight teacher model (SUTrack-B224~\cite{chen2024sutrack}) while operating at real-time speeds (54 FPS). Compared to other trackers (e.g., UnTrack~\cite{untrack}, SDSTrack~\cite{hou2024sdstrack}) and high-performance baselines (e.g., CAFormer~\cite{zhang2025cross}), our approach significantly dominates in terms of combined accuracy and efficiency.}
\vspace{-3mm}
\label{fig:teaser2}
\end{figure}

While multimodal trackers demonstrate promising gains, they introduce significant practical challenges, primarily regarding computational overhead and latency. Currently, multimodal tracking methods largely follow two paradigms. The first integrates a pretrained RGB tracker into a multimodal framework via visual prompting or dual-stream encoders~\cite{bat,hou2024sdstrack,vipt}. While effective, these architectures often struggle to balance deep multimodal fusion with inference speed. Visual prompting frequently suffers from insufficient fusion, whereas dual-stream designs incur heavy computational costs. To address this, unified frameworks like SUTrack have emerged, jointly pretraining on RGB and multimodal datasets to handle diverse modalities within a single model. While SUTrack pushes the boundaries of performance, its heavy architecture hinders deployment on resource-constrained edge devices. Consequently, developing a real-time, unified multimodal tracker remains a critical unmet need.

To visualize this dilemma, we present a comprehensive speed-accuracy comparison on the large-scale RGBT234 benchmark in Figure~\ref{fig:teaser2}. As illustrated, the current landscape is polarized: the teacher model, SUTrack-B224, establishes a high performance ceiling but suffers from sluggish inference, making it impractical for real-time applications. Conversely, existing efficient trackers like UnTrack (represented by blue bubbles) prioritize speed but at the cost of significant accuracy degradation, clustering in the lower-left region. Our proposed \textbf{DTrack} effectively breaks this stalemate. By occupying the top-right corner of the chart, DTrack achieves a real-time speed of 54 FPS—a $5\times$ speedup over the teacher, while maintaining state-of-the-art precision, significantly outperforming competitive baselines like CAFormer.

Achieving this "sweet spot" between efficiency and accuracy required us to rethink the standard model compression pipeline. A natural solution to bridge the gap between heavy, high-performance models (Teachers) and lightweight, efficient models (Students) is Knowledge Distillation(KD). By transferring learned representations from a teacher, KD allows compact models to approach the performance of their larger counterparts. However, we observe that even with KD, the efficiency gain is often capped by the underlying student architecture. Existing compression attempts in the tracking domain typically focus on backbone pruning while overlooking a critical source of redundancy: the prediction head. These heavy detection heads act as a computational bottleneck during inference. In this work, we argue that a holistic, efficient design must optimize both the feature extractor and the decision components. By strategically streamlining the prediction head channels, we successfully unlock further speed gains, pushing the inference frame rate from 42 FPS to 54 FPS.

Nevertheless, such aggressive architectural reduction introduces a new dilemma: the capacity gap. A lightweight student with a simplified head struggles to inherit the complex multimodal reasoning capabilities of the teacher using generic distillation objectives. To resolve this, we seek to answer a fundamental question: \textit{What matters most for distilling a unified multimodal tracker with a lightweight head?} Through a systematic analysis of 17 distillation strategies, we reveal that effective compression requires decoupling the knowledge transfer into two complementary streams. We observe that focusing distillation solely on one aspect, either spatial features or semantic logits, leads to a trade-off between localization precision and discrimination robustness. Based on this, we introduce a Dual-Alignment Distillation framework. First, we propose \textit{Spatial Structure Alignment} on the encoder's fused features to sharpen the student's focus on foreground targets ("\textit{Where to track}"). Second, we implement \textit{Semantic Distribution Alignment} via Logit-based distillation to align decision boundaries, transferring the "dark knowledge" of distractor relationships ("\textit{What to track}").

By synergizing these two streams, we achieve a minimalist yet highly effective strategy. Our contributions are summarized as follows:
\begin{itemize}
    \item We conduct the first comprehensive study of distillation mechanisms for unified multimodal tracking and propose a Dual-Alignment Distillation framework. This approach effectively compensates for the capacity gap of the lightweight head by decoupling feature-level spatial alignment from logit-level semantic transfer.
    
    \item We identify the prediction head as an overlooked bottleneck in efficient multimodal tracking. By designing a lightweight decoder, we significantly boost inference efficiency.

    \item Extensive experiments demonstrate that our method achieves state-of-the-art performance. As highlighted in Figure 1, our distilled model achieves 91.5\% MPR on RGBT234 and operates at 54 FPS on a single RTX 4090 GPU, representing a 5$\times$ speedup over the teacher model while maintaining superior accuracy.
\end{itemize}

\section{Related Work}

\subsection{Multimodal Tracking}

Multimodal object tracking models leverage additional modalities to augment RGB-based target tracking~\cite{hong2024onetracker,tan2025you}. ViPT was the first to unify different models under a single architecture; however, it requires modality-specific parameters to accommodate diverse downstream modalities~\cite{vipt}. Subsequent models have attempted to achieve unification, yet the inherent gaps between modalities often necessitate separate vision encoders to extract features from RGB and downstream modalities, incurring substantial computational overhead~\cite{hou2024sdstrack,untrack,tan2025xtrack,flextrack,sttrack}. While recent works have advanced modality-specific tracking using state space models~\cite{wang2025mambaevt} and specialized fusion mechanisms~\cite{zhu2024rgbt,xue2025fmtrack}, they still rely on disjointed architectures. To address this dilemma, SUTrack adopts a strategy of retraining a new model from scratch, enabling a single vision encoder to handle multiple modalities~\cite{chen2024sutrack}. Nevertheless, SUTrack's inference speed remains suboptimal on a single GPU. However, existing unified frameworks predominantly focus on accuracy improvements while overlooking the critical efficiency-accuracy trade-off required for edge deployment. Moreover, generic knowledge distillation techniques, while effective in single-modality settings, struggle to preserve the complex cross-modal reasoning patterns essential for robust multimodal tracking. This motivates our work to develop a task-aware compression framework that jointly optimizes architectural efficiency and knowledge transfer for real-time unified multimodal tracking. In this work, we conduct a comprehensive research of various approaches and employ model distillation to achieve real-time and accurate target tracking on consumer-grade GPU.

\subsection{Efficient Tracker}
Unlike multimodal tracking tasks, significant research effort has been devoted to developing lightweight algorithms for RGB-only visual tracking~\cite{zhu2023cross}. Early approaches typically followed the Siamese tracking pipeline and leveraged specially designed lightweight backbones to enable efficient real-time performance~\cite{yan2021lighttrack}. Subsequent work introduced more efficient architectures, such as a lightweight transformer-based feature fusion module~\cite{wei2024litetrack}. More recently, several methods have focused on optimizing Transformer-based trackers for deployment on edge devices. Examples include a feature sparsification strategy combined with a hierarchical cross-attention Transformer architecture to achieve real-time inference, an efficient Exemplar Transformer-based prediction head tailored for visual tracking, and a fully Transformer-based backbone enhanced through knowledge distillation and progressive depth pruning~\cite{mixformer,cui2023mixformerv2}. However, the majority of these lightweight trackers rely solely on RGB inputs and consequently struggle to maintain robust performance under challenging conditions. Although recent efforts explore lightweight multimodal adaptation~\cite{liu2024emtrack} and pruning-based distillation for single-modality tracking ~\cite{song2025exploring}. Moreover, the lightweighting of unified models remains largely unexplored.

\subsection{Knowledge Distillation}

Knowledge Distillation (KD) has emerged as a cornerstone paradigm for model compression, predominantly in image recognition, where it facilitates the transfer of dark knowledge through soft logits or intermediate feature representations~\cite{kd,dkd,pkd,rkd}. However, extending KD to unified multimodal tracking introduces unique challenges that transcend standard domain adaptation. Unlike classification, which primarily demands discriminative power over static categories, multimodal tracking necessitates the preservation of complex cross-modal reasoning capabilities—such as aligning thermal signatures with RGB textures or fusing asynchronous event streams with frame-based cues—capabilities that lightweight students often struggle to retain when distilled from heavy teachers~\cite{zhu2023cross}. Crucially, a systematic investigation tailored to the intricacies of unified multimodal tracking remains conspicuously absent. Current approaches predominantly adopt a naive transfer strategy, indiscriminately borrowing feature-centric alignment techniques from classification tasks while overlooking two critical aspects inherent to tracking architectures: first, the structural redundancy in prediction heads, which acts as a hidden computational bottleneck yet is rarely pruned or distilled in a task-aware manner; and second, the rich semantic guidance embedded in decision logits, which encodes valuable information about distractor relationships and decision boundaries often lost in coarse feature-level matching~\cite{mtkd,struckd,dsd}. Consequently, existing methods fail to disentangle spatial structural cues from semantic discriminative logic, leading to suboptimal accuracy-efficiency trade-offs where improvements in localization precision often come at the expense of discrimination robustness, or vice versa. Addressing this gap, our work poses a fundamental question: how can we effectively decouple and distill spatial and semantic knowledge to construct a streamlined yet robust unified tracker that maintains the teacher's comprehensive reasoning capabilities while achieving real-time inference?

\begin{figure*}[!t]
  \centering
  \includegraphics[width=0.95\linewidth,keepaspectratio]{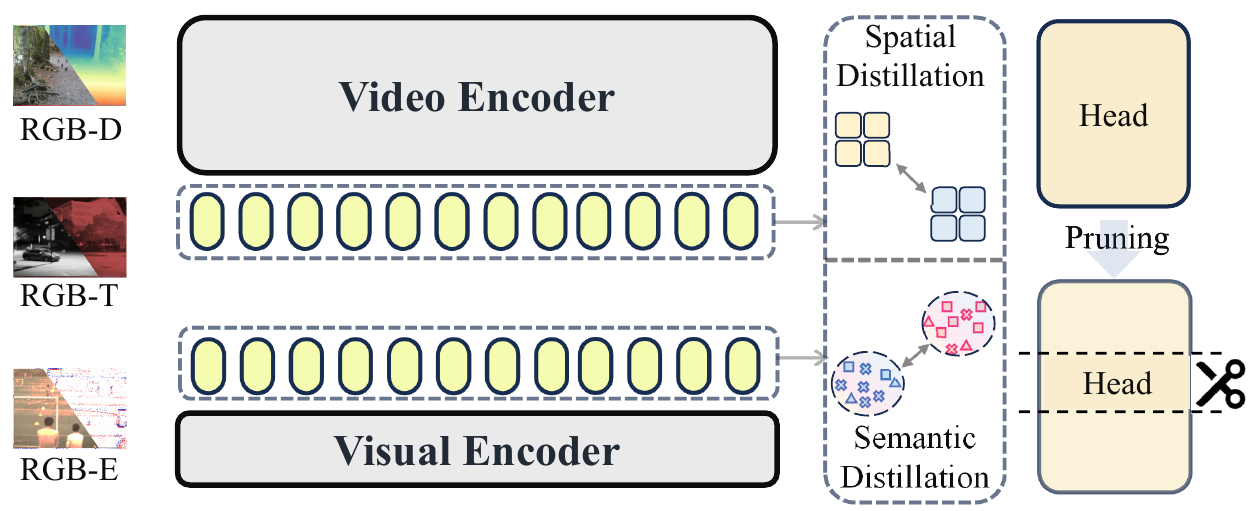}
  \caption{\textbf{Overview of our proposed \textit{DTrack} framework.} 
  The architecture is designed for unified multimodal tracking (handling RGB-D, RGB-E, and RGB-T inputs). 
We introduce a \textit{Lightweight Prediction Head} by structurally pruning the intermediate convolutional layers (visualized as dashed blocks) from the standard heavy decoder. This design significantly reduces computational redundancy, unlocking real-time inference speeds.
 To compensate for the capacity gap induced by pruning, we propose a \textit{Dual-Alignment Distillation} strategy on the encoder features. This framework decouples knowledge transfer into two complementary streams: \textit{Spatial Structure Alignment} (sharpening localization focus) and \textit{Semantic Distribution Alignment} (transferring discriminative logic), ensuring the compact student effectively mimics the robust teacher.}
  \label{fig:framework}
\end{figure*}

\section{Method}

\subsection{Dual-Alignment Distillation Framework}
\label{sec:distillation}
Directly applying generic distillation objectives fails to balance the dual requirements of tracking: precise localization and robust discrimination. To resolve this, we propose a \textbf{Dual-Alignment Distillation} strategy that disentangles knowledge transfer into two complementary streams: \textit{Spatial Structure Alignment} (focusing on "Where") and \textit{Semantic Distribution Alignment} (focusing on "What").

Let $\mathbf{F}_s, \mathbf{F}_t \in \mathbb{R}^{B \times C \times H \times W}$ denote the high-level feature maps from the last layer of the student and teacher encoders, respectively.

\subsubsection{Spatial Structure Alignment}
The first stream aims to sharpen the student's focus on the foreground target, which is critical for precise localization. However, a direct element-wise matching of feature maps is suboptimal because the teacher and student often exhibit significant discrepancies in feature magnitude due to their architectural differences (e.g., channel capacity).

To address this, we propose to align the \textit{spatial structure} rather than absolute magnitudes. We project the feature vectors onto a unit hypersphere by applying $L_2$ normalization along the channel dimension. This operation decouples the direction of the feature vector (which encodes structural attention) from its norm. Let $\mathbf{f}^{(i)}$ denote the feature vector at the $i$-th spatial position. The normalized features $\hat{\mathbf{F}}$ are computed as:
\begin{equation}
    \hat{\mathbf{F}}_s = \frac{\mathbf{F}_s}{\|\mathbf{F}_s\|_2}, \quad \hat{\mathbf{F}}_t = \frac{\mathbf{F}_t}{\|\mathbf{F}_t\|_2},
\end{equation}
where $\|\cdot\|_2$ denotes the $L_2$ norm. We then minimize the Mean Squared Error (MSE) between these normalized representations:
\begin{equation}
    \mathcal{L}_{spatial} = \frac{1}{N} \sum_{i=1}^{N} \left\| \hat{\mathbf{F}}_s^{(i)} - \hat{\mathbf{F}}_t^{(i)} \right\|^2_2,
\end{equation}
where $N = B \times H \times W$. By enforcing consistency on the unit hypersphere, we ensure that the student mimics the teacher's directional attention towards the target object, regardless of the intensity of activations.

\subsubsection{Semantic Distribution Alignment}
While the spatial stream ensures geometric precision, it does not explicitly constrain the statistical distribution of activations, which encodes the semantic confidence and inter-class relationships (e.g., distinguishing the target from similar distractors). To transfer this "dark knowledge," we introduce a semantic alignment stream.

Since the raw activation distributions of the teacher and student may lie in different domains, directly aligning them can be unstable. Therefore, we first standardize the features to a unified distribution space. We compute the channel-wise mean $\boldsymbol{\mu}$ and standard deviation $\boldsymbol{\sigma}$ for each sample:
\begin{equation}
    \boldsymbol{\mu} = \frac{1}{C} \sum_{c=1}^{C} \mathbf{F}_{c}, \quad \boldsymbol{\sigma} = \sqrt{\frac{1}{C} \sum_{c=1}^{C} (\mathbf{F}_{c} - \boldsymbol{\mu})^2}.
\end{equation}
The standardized semantic map $\tilde{\mathbf{Z}}$ is obtained by:
\begin{equation}
    \tilde{\mathbf{Z}} = \frac{\mathbf{F} - \boldsymbol{\mu}}{\boldsymbol{\sigma} + \epsilon},
\end{equation}
where $\epsilon=10^{-6}$ ensures numerical stability. This standardization effectively filters out instance-specific magnitude variations, leaving purely relative semantic information. We then apply a temperature scaling parameter $T$ to soften the distribution and employ the Kullback-Leibler (KL) Divergence to align the student's semantic probability with the teacher's:
\begin{equation}
    \mathcal{L}_{semantic} = T^2 \cdot \text{KL}\left( \text{softmax}\left(\frac{\tilde{\mathbf{Z}}_t}{T}\right) \Big\| \text{softmax}\left(\frac{\tilde{\mathbf{Z}}_s}{T}\right) \right).
\end{equation}
Here, the $\text{softmax}(\cdot)$ is applied over the spatial dimensions to represent the probability distribution of semantic confidence. This objective transfers the teacher's robust discriminative logic to the student.

\subsubsection{Overall Optimization}
The final training objective synergizes the task-specific tracking loss $\mathcal{L}_{base}$ with our dual-alignment distillation terms:
\begin{equation}
    \mathcal{L}_{total} = \mathcal{L}_{base} + \lambda_{spa} \mathcal{L}_{spatial} + \lambda_{sem} \mathcal{L}_{semantic},
\end{equation}
where $\lambda_{spa}$ and $\lambda_{sem}$ are balancing hyperparameters. This coupled formulation ensures that the lightweight student learns both \textit{where to look} (via spatial alignment) and \textit{what to distinguish} (via semantic alignment), effectively compensating for the capacity loss induced by head pruning.

\subsection{Lightweight Prediction Head Pruning}
\label{sec:head_pruning}
Unlike classification tasks, the prediction heads in object tracking are notoriously heavy. Prior works on efficient tracking have predominantly focused on backbone pruning, inadvertently neglecting the substantial overhead introduced by the prediction head. We challenge this convention by targeting the prediction head as a critical, yet overlooked, computational bottleneck. Conventional multimodal trackers utilize deep, stacked convolutional layers to transform high-level fused features into task-specific outputs. Although this design ensures non-linear capacity, it incurs significant latency that hinders real-time edge deployment. We argue that because the encoder's fused features are already semantically robust, the deep mapping within the decoder contains structural redundancy that can be safely eliminated.

Formally, let $\mathbf{X} \in \mathbb{R}^{C \times H \times W}$ denote the input feature map from the encoder. A conventional prediction branch consists of a sequence of convolutional blocks. Let $\phi(\cdot)$ represent the non-linear activation function (specifically, $\phi(\mathbf{x}) = \text{ReLU}(\text{BN}(\mathbf{x}))$). The operation of a standard 5-layer branch is formulated as:
\begin{align}
    \mathbf{H}_1 &= \phi(\mathbf{W}_1 * \mathbf{X}), \notag \\
    \mathbf{H}_2 &= \phi(\mathbf{W}_2 * \mathbf{H}_1), \notag \\
    \mathbf{H}_3 &= \phi(\mathbf{W}_3 * \mathbf{H}_2), \notag \\
    \mathbf{H}_4 &= \phi(\mathbf{W}_4 * \mathbf{H}_3), \notag \\
    \mathbf{S}_{orig} &= \mathbf{W}_5 * \mathbf{H}_4,
\end{align}
where $*$ denotes the convolution operation, $\mathbf{W}_i$ represents the weights of the $i$-th layer, and $\mathbf{S}_{orig}$ is the final output map.

To streamline the inference process, we perform structural pruning by eliminating intermediate layers (specifically layers 2 and 4). This reduces the depth of the network while retaining the essential feature transformation capabilities. The pruned lightweight branch is reformulated as:
\begin{align}
    \mathbf{H}_1 &= \phi(\mathbf{W}_1 * \mathbf{X}), \notag \\
    \mathbf{H}_3 &= \phi(\mathbf{W}_3 * \mathbf{H}_1), \notag \\
    \mathbf{S}_{light} &= \mathbf{W}_5 * \mathbf{H}_3.
\end{align}
This reduction is applied uniformly across all three task-specific branches: the center classification branch ($\mathbf{S}_{ctr}$), the offset regression branch ($\mathbf{S}_{off}$), and the size regression branch ($\mathbf{S}_{size}$). 

By reducing the branch depth from 5 to 3, we significantly decrease the floating-point operations (FLOPs) and memory access costs, pushing the inference speed from 42 FPS to 54 FPS. 

\begin{table}[!t]\normalsize
    \caption{Efficiency analysis of different model variants. We report parameter count, FLOPs, and inference speed (FPS).}
\label{tab:efficiency}
\vspace{-2mm}
  \centering
\resizebox{\linewidth}{!}{
  \setlength{\tabcolsep}{4mm}{
    \small
    \begin{tabular}{l|ccc}
    \toprule
    Model & Params (M)  & FLOPs (G) & FPS \\
    \midrule[0.5pt]
        SUTrack-B (Teacher) & 70.0 & 23.0  & 11 \\
        Ours (w/o Head Pruning) & 22.0 & 6.0  & 47 \\
        Ours  & \textbf{18.0} & \textbf{4.0}  & \textbf{54} \\
    \bottomrule
    \end{tabular}
    }
  }
  \vspace{-4mm}
\end{table}

\begin{table}[!t]\normalsize
    \caption{SOTA comparisons on RGB-Event tracking dataset.
    }
\label{tab-sota-rgbe}
  \centering
\resizebox{1\linewidth}{!}{
  \setlength{\tabcolsep}{6mm}{
    \small
    \begin{tabular}{l|cc}
    \toprule
    \multirow{2}*{Method} & \multicolumn{2}{c}{VisEvent}\\
        \cline{2-3}
 & PR& SR\\
    \midrule[0.5pt]

        DTrack (Ours) &\textbf{\textcolor{red}{79.7}} &\textbf{\textcolor{red}{62.3}}\\

        \midrule[0.1pt]


OneTracker~\cite{hong2024onetracker}&76.7 &60.8\\

SDSTrack~\cite{hou2024sdstrack} &76.7 &59.7\\

UnTrack~\cite{untrack}&75.5 &58.9 \\
ViPT~\cite{vipt}&75.8 &59.2\\
 
ProTrack~\cite{protrack}&63.2 &47.1\\

OSTrack~\cite{ostrack} &69.5&53.4 \\
SiamRCNN\_E~\cite{SiamRCNN} &65.9 &49.9 \\
TransT\_E~\cite{transt} &65.0 &47.4 \\
LTMU\_E~\cite{LTMU} &65.5&45.9 \\
PrDiMP\_E~\cite{PrDiMP} &64.4 &45.3 \\
VITAL\_E~\cite{VITAL} &64.9&41.5 \\
MDNet\_E~\cite{MDNet} &66.1&42.6 \\
ATOM\_E~\cite{atom} &60.8 &41.2 \\
SiamCar\_E~\cite{Stark} &59.9 &42.0 \\
SiamBAN\_E~\cite{siamban} &59.1 &40.5 \\
SiamMask\_E~\cite{SiamMask} &56.2 &36.9 \\

    \bottomrule
    \end{tabular}
    }
  }
  \label{rgbe}
\end{table}

\begin{table}[!t]
    \caption{SOTA comparisons on RGB-Thermal tracking datasets.
    }

\label{tab-sota-rgbt}
  \centering
\resizebox{1\linewidth}{!}{
  \setlength{\tabcolsep}{2mm}{
    \small
    \begin{tabular}{l|ccccc}
    \toprule
    \multirow{2}*{Method} & \multicolumn{2}{c}{LasHeR} & & \multicolumn{2}{c}{RGBT234} \\
        \cline{2-3} \cline{5-6}
 & PR & SR & &MPR &MSR \\
    \midrule[0.5pt]

        DTrack (Ours) &\textbf{\textcolor{red}{72.4}}&\textbf{\textcolor{red}{57.6}} & &\textcolor{red}{\textbf{91.5}} &\textcolor{red}{\textbf{67.0}}\\
        \midrule[0.1pt]

OneTracker~\cite{hong2024onetracker} &67.2 &53.8 & & 85.7 &64.2\\
SDSTrack~\cite{hou2024sdstrack} &66.5&53.1 & &84.8 &62.5\\
UnTrack~\cite{untrack} &64.6&51.3 & &84.2& 62.5\\
ViPT~\cite{vipt} &65.1 &52.5 & &83.5&61.7\\
ProTrack~\cite{protrack} &53.8&42.0 & & 79.5& 59.9\\
OSTrack~\cite{ostrack} &51.5 &41.2 & &72.9&54.9 \\
APFNet~\cite{apfnet} &50.0&36.2 &&82.7 &57.9 \\
CMPP~\cite{cmpp} &-&- &&82.3 &57.5 \\
JMMAC~\cite{jmmac} &-&- &&79.0 &57.3 \\
CAT~\cite{cat} &45.0&31.4 &&80.4 &56.1 \\
FANet~\cite{fanet} &44.1&30.9 & &78.7&55.3 \\
DAPNet~\cite{dapnet} &43.1&31.4 & &-&- \\
DAFNet~\cite{dafnet} &-&- &&79.6 &54.4 \\
MaCNet~\cite{macnet} &-&- &&79.0 &55.4 \\

    \bottomrule
    \end{tabular}
    }
  }
  \label{tab:rgbt}
\end{table}

\section{Experiments}

\subsection{Implementation Details}
Consistent with recent trackers~\cite{sttrack, hong2024onetracker, hou2024sdstrack}, our tracker is initialized with a pretrained encoder~\cite{mcitrack,fastitpn} trained on large-scale datasets~\cite{got10k, trackingnet, coco, lasot}. The model is trained on 2×4090 GPUs with a batch size of 16 using the AdamW optimizer, with a base learning rate of 3e-4. Training consists of 150 epochs, each containing 600,000 image pairs. 
For RGB-Thermal, we use LasHeR~\cite{li2021lasher} for training, VisEvent ~\cite{wang2023visevent} for RGB-Event, and DepthTrack~\cite{depthtrack} for RGB-Depth. To create a unified framework, we jointly train these three datasets in a single training process.
During inference, we disable masking operations to fully leverage all available modalities while retaining the robustness learned through training phase masking simulations.

\subsection{Multi-Angle Efficiency Analysis.}
To achieve real-time inference without compromising accuracy, we systematically reduce computational redundancy from two complementary perspectives:

\noindent\textbf{(1) Backbone Distillation.} By transferring knowledge from the heavy teacher (SUTrack-B) to a lightweight student architecture, we reduce the overall parameter count from 70.0M to 22.0M (68.6\% reduction) and FLOPs from 23.0G to 6.0G (73.9\% reduction), while preserving the encoder's semantic representation capability through our dual-alignment distillation framework.

\noindent\textbf{(2) Prediction Head Pruning.} Beyond backbone compression, we identify the prediction head as an overlooked efficiency bottleneck. By structurally pruning intermediate convolutional layers (reducing depth from 5 to 3), we further decrease head-specific FLOPs from 2.1G to 0.8G (61.9\% reduction) with negligible accuracy loss, pushing inference speed from 47 FPS to 54 FPS.

As summarized in Table~\ref{tab:efficiency}, our holistic optimization strategy yields a \textbf{4.9× end-to-end speedup} over the teacher model (11 FPS → 54 FPS) with only 25.7\% of the original parameters, demonstrating that thoughtful architectural compression can effectively bridge the gap between accuracy and efficiency for edge deployment.
Notably, all efficiency comparisons are conducted under an identical hardware configuration. Unlike approaches that report optimistic peak performance by selecting the maximum FPS across multiple trials or employing aggressive GPU warm-up strategies, we report the average inference FPS measured over standard runs. This protocol ensures a fair and realistic assessment of actual deployment capabilities.

\subsection{Main Results}

To validate the effectiveness and robustness of our proposed method, we conduct a comprehensive evaluation across three distinct multimodal tracking tasks: RGB-Event, RGB-Thermal, and RGB-Depth. Detailed comparisons with state-of-the-art (SOTA) methods are presented below.

\begin{table}[!t]
    \caption{SOTA comparisons on RGB-Depth tracking datasets.
    }
\label{tab-sota-rgbd}
  \centering
\resizebox{1\linewidth}{!}{
  \setlength{\tabcolsep}{1.5mm}{
    \small
    \begin{tabular}{l|ccc c ccc}
    \toprule
    \multirow{2}*{Method} & \multicolumn{3}{c}{DepthTrack} & & \multicolumn{3}{c}{VOT-RGBD22} \\
        \cline{2-4} \cline{6-8}
 & F-score & RE & PR& & EAO & Acc.& Rob. \\
    \midrule[0.5pt]
DTrack (Ours) &57.1&56.9&57.0 & &\textcolor{red}{\textbf{74.7}}&\textbf{\textcolor{red}{82.0}} &\textcolor{red}{\textbf{90.5}}\\     
\midrule[0.1pt]
OneTracker~\cite{hong2024onetracker} &60.9 &60.4 &60.7 & &72.7 & 81.9 & 87.2\\
SDSTrack~\cite{hou2024sdstrack} &61.4 &60.9 & 61.9 & &72.8 & 81.2 & 88.3\\
UnTrack~\cite{untrack} &61.0&60.8&61.1 & &72.1&82.0 &86.9\\
ViPT~\cite{vipt} &59.4&59.6&59.2 & &72.1&81.5 &87.1\\
ProTrack~\cite{protrack} &57.8&57.3&58.3 & &65.1&80.1&80.2\\
SPT~\cite{rgbd1k} &53.8&54.9&52.7 & &65.1&79.8&85.1\\
SBT-RGBD~\cite{sbt} &-&-&- & &70.8&80.9&86.4\\
OSTrack~\cite{ostrack} &52.9&52.2&53.6 & &67.6&80.3&83.3\\
DeT~\cite{depthtrack} &53.2&50.6&56.0 & &65.7&76.0&84.5\\
DMTrack~\cite{vot2022} &-&-&- & &65.8&75.8&85.1\\
DDiMP~\cite{vot2020} &48.5&56.9&50.3 & &-&-&-\\
ATCAIS~\cite{vot2020} &47.6&45.5&50.0 & &55.9&76.1&73.9\\
LTMU-B~\cite{LTMU} &46.0&41.7&51.2 & &-&-&-\\
GLGS-D~\cite{vot2020} &45.3&36.9&58.4 & &-&-&-\\
DAL~\cite{dal} &42.9&36.9&51.2 & &-&-&-\\
LTDSEd~\cite{VOT2019} &40.5&38.2&43.0 & &-&-&-\\
Siam-LTD~\cite{vot2020} &37.6&34.2&41.8 & &-&-&-\\
SiamM-Ds~\cite{VOT2019} &33.6&26.4&46.3 & &-&-&-\\
CA3DMS~\cite{ca3dms} &22.3&22.8&21.8 & &-&-&-\\
DiMP~\cite{dimp} &-&-&- & &54.3&70.3&73.1\\
ATOM~\cite{atom} &-&-&- & &50.5&59.8&68.8\\
    \bottomrule
    \end{tabular}
    }
  }
\label{tab:rgbd}
\end{table}

\vspace{1mm} \noindent \textbf{RGB-E Tracking Results}

For RGB-Event tracking, we utilize the VisEvent dataset~\cite{wang2023visevent}, which originally contains 820 RGB-E video pairs. Following the standard protocol, we filter out sequences with missing event data or timestamp misalignment, resulting in a curated set of 377 training sequences and 172 testing sequences. We employ Precision Rate (PR) and Success Rate (SR) as the primary evaluation metrics.

\textbf{Quantitative Analysis:} As shown in Table~\ref{tab-sota-rgbe}, our tracker achieves state-of-the-art performance on VisEvent. Specifically, it obtains a Precision score of 79.7\% and an AUC score of 62.3\%. Compared to the leading competitor OneTracker~\cite{hong2024onetracker}, our method delivers a substantial improvement of 3.0\% in Precision and 1.5\% in AUC. This demonstrates the superiority of our distilled student model in fusing asynchronous event streams with RGB data, effectively surpassing heavy-weight baselines like SDSTrack~\cite{hou2024sdstrack} and ViPT~\cite{vipt}.

\vspace{2mm} \noindent \textbf{RGB-T Tracking Results}

We evaluate our method on two large-scale benchmarks: RGBT234 and LasHeR. RGBT234 comprises 234 RGB-T video pairs with 12 attribute annotations, providing a diverse testbed for thermal tracking. LasHeR is currently the largest dataset in this domain, containing 1,244 sequences with over 730,000 frame pairs, split into a training set of 979 videos and a testing set of 245 videos. Performance is evaluated using Maximum Precision Rate (MPR) and Maximum Success Rate (MSR).

\textbf{Quantitative Analysis:} Table~\ref{tab-sota-rgbt} illustrates the comparison results. On the challenging LasHeR dataset, our method demonstrates dominance, achieving 72.4\% in Precision and 57.6\% in Success Rate. Notably, it outperforms the second-best method, OneTracker, by a significant margin of 5.2\% in Precision. Similarly, on RGBT234, our tracker sets a new record with 91.5\% MPR and 67.0\% MSR, surpassing OneTracker by 5.8\% and 2.8\%, respectively. These results confirm that our distillation framework effectively retains the discriminative power of the teacher model while handling the thermal modality's challenges, such as cross-modality misalignment and thermal crossover.

\vspace{2mm} \noindent \textbf{RGB-D Tracking Results}

For RGB-Depth tracking, evaluations are conducted on DepthTrack and VOT-RGBD22. DepthTrack is a comprehensive benchmark consisting of 150 training and 50 testing sequences, designed to address challenging scenarios like deformable objects and clutter. The metrics include Precision (PR), Recall (R), and F-score (F). VOT-RGBD22 is evaluated using the Expected Average Overlap (EAO), Accuracy (Acc.), and Robustness (Rob.).

\textbf{Quantitative Analysis:} The results are reported in Table~\ref{tab-sota-rgbd}. On the VOT-RGBD22 benchmark, our method achieves the best overall performance with an EAO of 74.7\%, an Accuracy of 82.0\%, and a Robustness score of 90.5\%. It surpasses the robust SDSTrack by 1.9\% in EAO, highlighting its effectiveness in depth-aware tracking. On the DepthTrack dataset, our lightweight model maintains competitive performance with an F-score of 57.1\%, balancing efficiency and accuracy against heavy-weight counterparts.

\begin{table}[!t]\normalsize
    \caption{Ablation study on the depth of the distilled feature layer (RGBT234 dataset). Distilling deeper layers generally yields better performance, with the final layers (Layer 10-12) providing the most robust representations.
    }
\label{tab:interfeat}
\vspace{-2mm}
  \centering
\resizebox{0.8\linewidth}{!}{
  \setlength{\tabcolsep}{5mm}{
    \small
    \begin{tabular}{l|cc}
    \toprule
    \multirow{2}*{\textbf{Distillation Stage}} & \multicolumn{2}{c}{\textbf{RGBT234}} \\
        \cline{2-3}
 & MPR & MSR \\
    \midrule[0.5pt]
        Encoder Layer 2 & 87.1 & 64.3 \\
        Encoder Layer 4 & 88.1 & 64.7 \\
        Encoder Layer 6 & 89.0 & 66.0 \\
        Encoder Layer 8 & 89.7 & 66.9 \\
        Encoder Layer 10 & 89.8 & \textbf{67.0} \\
        Encoder Layer 12 (Final) & \textbf{90.1} & 66.7 \\
    \bottomrule
    \end{tabular}
    }
  }
  \vspace{-4mm}
\end{table}

\section{Ablation Studies and Discussion}

\begin{table*}[!t]
  \caption{Comprehensive comparison with state-of-the-art methods and various distillation strategies across five multimodal tracking benchmarks. Distillation methods are \textbf{re-ordered and labeled} by alignment type (Spatial-only vs. Semantic-only). \textbf{DTrack} is the only Dual-Alignment method. The best results among all distillation methods are highlighted in \textbf{\textcolor{red}{red}}.}
  \label{tab:all_datasets}
  \centering
  \resizebox{\textwidth}{!}{
    \setlength{\tabcolsep}{1.1mm}{
      \begin{tabular}{l | c | cc | cc | cc | ccc | ccc | c}
        \toprule
        \multirow{2}{*}{\textbf{Method}} & \multirow{2}{*}{\textbf{Type}} & \multicolumn{2}{c|}{\textbf{VisEvent}} & \multicolumn{2}{c|}{\textbf{LasHeR}} & \multicolumn{2}{c|}{\textbf{RGBT234}} & \multicolumn{3}{c|}{\textbf{DepthTrack}} & \multicolumn{3}{c|}{\textbf{VOT-RGBD22}} & \multirow{2}{*}{\textbf{FPS}} \\
        \cline{3-14}
        & & PR & SR & PR & SR & MPR & MSR & RE & PR & F-score & EAO & Acc. & Rob. & \\
        \midrule
        \multicolumn{15}{c}{\textit{State-of-the-Art Baselines}} \\
        \midrule
        SDSTrack & - & 76.7 & 59.7 & 66.5 & 53.1 & 84.8 & 62.5 & 60.9 & 61.9 & 61.4 & 72.8 & 81.2 & 88.3 & 28 \\
        UnTrack & - & 75.5 & 58.9 & 63.7 & 51.3 & 84.2 & 62.5 & 61.0 & 61.0 & 61.0 & 71.8 & 82.0 & 86.4 & 14 \\
        XTrack-B & - & 77.5 & 60.9 & 69.1 & 55.7 & 87.4 & 64.9 & 62.0 & 61.5 & 61.8 & 73.4 & 81.5 & 88.6 & 8 \\
        CAFormer & - & - & - & 70.0 & 55.6 & 88.3 & 66.4 & - & - & - & - & - & - & 42 \\
        SUTrack-B224 (Teacher) & - & 79.9 & 62.7 & 74.5 & 59.9 & 92.2 & 69.5 & 65.7 & 64.5 & 65.1 & 76.5 & 82.8 & 91.8 & 11 \\
        \midrule
        \multicolumn{15}{c}{\textit{Distillation Comparisons}} \\
        \midrule
        \midrule
        CWD~\cite{cwd} & \textit{Spatial} & 77.9 & 60.4 & 68.7 & 55.0 & 88.8 & 65.7 & 57.0 & 56.0 & 56.5 & 71.6 & 81.0 & 87.5 & 54 \\
        L2 & \textit{Spatial} & 78.2 & 60.6 & 70.9 & 54.9 & 87.3 & 64.8 & 57.4 & 57.1 & 57.3 & 72.1 & 81.4 & 88.1 & 54 \\
        FNKD~\cite{fnkd} & \textit{Spatial} & 79.2 & 61.8 & 71.9 & 57.5 & 89.2 & 66.3 & 59.4 & 59.1 & 59.2 & 73.9 & 81.3 & 89.9 & 54 \\
        SP~\cite{sp} & \textit{Spatial} & 77.6 & 59.9 & 70.9 & 56.9 & 87.6 & 64.5 & 57.1 & 56.1 & 56.6 & 71.1 & 81.2 & 86.7 & 54 \\
        CUSTOMKD~\cite{CustomKD} & \textit{Spatial} & 77.0 & 59.7 & 68.4 & 54.8 & 88.0 & 64.7 & 59.7 & 59.1 & 59.4 & 72.1 & 81.2 & 88.2 & 54 \\
        CBAM~\cite{cbam} & \textit{Spatial} & 77.4 & 60.0 & 68.6 & 55.1 & 87.1 & 64.0 & 56.7 & 56.4 & 56.6 & 72.0 & 80.9 & 88.2 & 54 \\
        CAD~\cite{cad} & \textit{Spatial} & 79.6 & 62.2 & 70.8 & 56.7 & 89.5 & 65.8 & 58.1 & 57.3 & 57.7 & 74.6 & 81.8 & 90.4 & 54 \\
        StrucKD~\cite{struckd} & \textit{Spatial} & 77.9 & 60.2 & 67.0 & 53.6 & 86.9 & 64.4 & 54.9 & 55.0 & 54.9 & 71.5 & 80.6 & 87.7 & 54 \\
        DSD~\cite{dsd} & \textit{Spatial} & 77.4 & 60.0 & 69.0 & 55.3 & 88.3 & 65.7 & 56.6 & 56.3 & 56.4 & 72.1 & 81.3 & 87.9 & 54 \\
        \midrule
        KD~\cite{kd} & \textit{Semantic} & 78.5 & 60.8 & 68.9 & 55.0 & 86.7 & 64.3 & 57.4 & 57.1 & 57.3 & 72.3 & 81.3 & 88.3 & 54 \\
        DHO~\cite{dho} & \textit{Semantic} & 76.4 & 59.5 & 68.5 & 54.9 & 87.2 & 64.2 & 56.5 & 55.6 & 56.1 & 72.6 & 80.8 & 89.1 & 54 \\
        DKD~\cite{dkd} & \textit{Semantic} & 78.5 & 61.0 & 69.8 & 55.8 & 88.7 & 65.8 & 57.4 & 56.5 & 56.9 & 71.6 & 81.0 & 88.1 & 54 \\
        GAN~\cite{gans} & \textit{Semantic} & 77.3 & 59.9 & 67.7 & 54.2 & 86.5 & 64.5 & 58.0 & 57.3 & 57.7 & 72.2 & 81.2 & 88.2 & 54 \\
        LOGIT~\cite{kd} & \textit{Semantic} & 79.4 & 62.1 & 72.3 & 57.7 & 90.8 & 66.6 & 58.9 & 58.0 & 58.4 & 74.6 & 81.9 & 90.4 & 54 \\
        MTKD~\cite{mtkd} & \textit{Semantic} & 74.4 & 57.6 & 63.1 & 50.8 & 83.4 & 61.9 & 50.2 & 57.7 & 53.7 & 68.0 & 80.3 & 83.4 & 54 \\
        RKD~\cite{rkd} & \textit{Semantic} & 78.4 & 60.8 & 69.0 & 55.4 & 87.6 & 65.0 & 59.0 & 58.4 & 58.7 & 71.9 & 80.8 & 88.1 & 54 \\
        \midrule
        \textbf{DTrack (Ours)} & \textbf{\textcolor{blue}{Dual-Alignment}} & \textbf{\textcolor{red}{79.7}} & \textbf{\textcolor{red}{62.3}} & \textbf{\textcolor{red}{72.4}} & 57.6 & \textbf{\textcolor{red}{91.5}} & \textbf{\textcolor{red}{67.0}} & 57.1 & 56.9 & 57.0 & \textbf{\textcolor{red}{74.7}} & \textbf{\textcolor{red}{82.0}} & \textbf{\textcolor{red}{90.5}} & 54 \\
        \bottomrule
      \end{tabular}
    }
  }
\end{table*}


\begin{figure}[!t]
  \centering
  \includegraphics[width=0.7\linewidth]{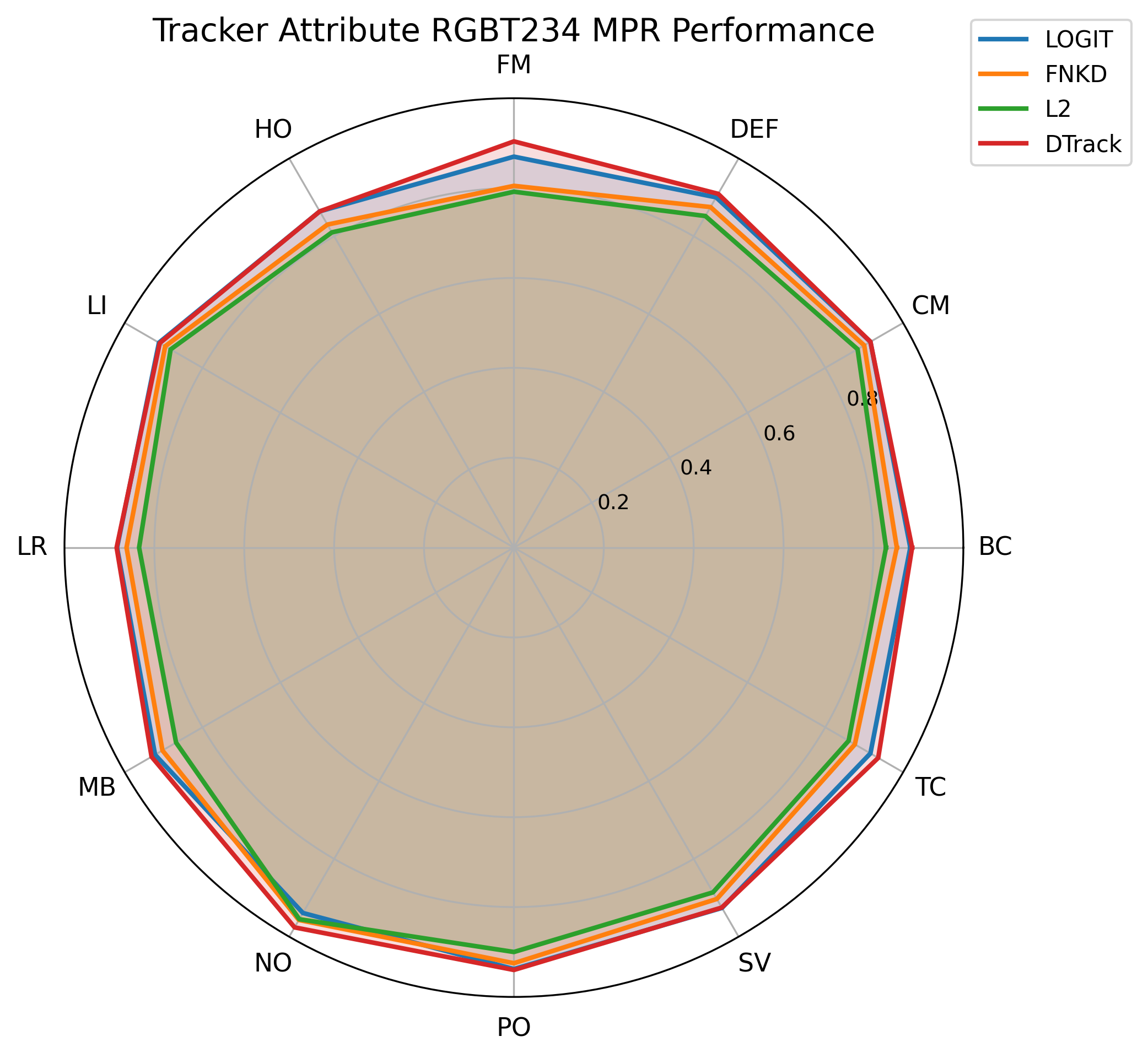}
  \caption{\textbf{Attribute-based comparison on the RGBT234 dataset (MPR).} The radar chart illustrates the performance of different distillation strategies across 12 challenging attributes (e.g., Fast Motion, Occlusion, Low Illumination). Our proposed DTrack consistently encompasses other methods, demonstrating superior robustness across diverse tracking scenarios.}
  \label{fig:radar}
\end{figure}

\noindent \textbf{Impact of Intermediate Feature Distillation.} 
To pinpoint the optimal abstraction level for knowledge transfer, we conducted a layer-wise ablation on the RGBT234 benchmark by varying the distillation target across the encoder's depth (from \texttt{Layer2} to \texttt{Layer12}). As detailed in Table~\ref{tab:interfeat}, we observe a distinct positive correlation between feature depth and tracking proficiency. Specifically, advancing the distillation target from the shallow \texttt{Layer2} to the deep \texttt{Layer12} yields substantial gains, boosting MPR from 87.1\% to 90.1\% and MSR from 64.3\% to 66.7\%. This trajectory suggests that the lightweight student benefits significantly more from high-level semantic representations—which encapsulate robust cross-modal logic—than from low-level structural cues. While the MSR plateaus and peaks slightly earlier at \texttt{Layer10} (67.0\%), the final layer (\texttt{Layer12}) delivers the highest precision. Crucially, this performance scaling incurs no latency cost, maintaining a constant 54 FPS across all configurations. Consequently, we adopt the final encoder layer as the default distillation target to maximize the precision-robustness synergy.

\noindent \textbf{Analysis on Distillation Mechanisms.}
As quantitatively presented in Table~\ref{tab:all_datasets}, we conduct the first systematic study of 17 distillation strategies for unified multimodal tracking. The table is deliberately grouped into Spatial-only and Semantic-only categories to highlight a fundamental limitation of existing paradigms: no single-stream approach can simultaneously achieve strong localization precision (“Where to track”) and robust discriminative power (“What to track”).
In the Spatial-only group (CWD, L2, FNKD, SP, CUSTOMKD, CBAM, CAD, StrucKD, DSD), methods excel at transferring structural attention on encoder features, yielding competitive precision on simpler attributes. However, they consistently falter in discrimination robustness. For example, the strongest Spatial-only variant (CAD) reaches 89.5\% MPR on RGBT234 and 74.6 EAO on VOT-RGBD22, yet lags behind in challenging cross-modal scenarios.
Conversely, the Semantic-only group (KD, DHO, DKD, GAN, LOGIT, MTKD, RKD) focuses on logit-level decision boundaries and dark knowledge, delivering superior discrimination. LOGIT, for instance, achieves the highest Semantic-only MPR of 90.8\% on RGBT234 and 74.6 EAO on VOT-RGBD22. Nevertheless, these methods suffer from suboptimal localization (e.g., lower PR on LasHeR and VisEvent compared to top Spatial-only counterparts).
Our Dual-Alignment Distillation (DTrack) is the only method that breaks this trade-off by explicitly decoupling and synergizing the two streams. On RGBT234, DTrack reaches 91.5\% MPR — surpassing the best Spatial-only result (89.5\%) by 2.0\% and the best Semantic-only result (90.8\%) by 0.7\%. Similar gains appear across all five benchmarks: it ranks first among all distillation methods on VisEvent (79.7 PR / 62.3 SR), LasHeR (72.4 PR / 57.6 SR), RGBT234 (91.5 MPR / 67.0 MSR), and VOT-RGBD22 (74.7 EAO / 82.0 Acc. / 90.5 Rob.). Crucially, as shown in our component ablation (Table~\ref{tab:ablation_rgbt}), simply adding Spatial-Only and Semantic-Only still cannot match the joint performance, confirming that the two alignments are mutually reinforcing rather than redundant.
By decoupling spatial structure alignment from semantic distribution alignment, DTrack successfully condenses the teacher’s complex cross-modal reasoning into a lightweight student while maintaining real-time speed of 54 FPS. This consistent state-of-the-art performance across RGB-E, RGB-T, and RGB-D modalities validates the necessity and effectiveness of our task-aware Dual-Alignment framework.

\noindent \textbf{Generalization Analysis on RGBD1K.} While our method achieves competitive results on DepthTrack, we observe a performance gap compared to the teacher model due to the specific domain challenges of that benchmark. To further validate the generalization capability and robustness of our distilled tracker across different scenarios, we extend our evaluation to the RGBD1K dataset, which contains 1,000 sequences and offers a broader diversity of scenes. As reported in Table~\ref{tab:rgbd1k_sota}, our method demonstrates exceptional generalization. Unlike the performance drop observed on DepthTrack, our lightweight student model achieves an F-score of 54.9\% on RGBD1K, fully matching the performance of the heavy teacher model (SUTrack-B224). More notably, in terms of PR, our method reaches 57.3\%, surpassing the teacher's 56.0\% by 1.3\%. This result confirms that our coupled distillation strategy effectively transfers the discriminative power of the teacher without overfitting to specific dataset distributions. Furthermore, compared to other efficient trackers like UnTrack (F-score 51.3\%) and SDSTrack (F-score 49.3\%), our method maintains a significant lead while running at a real-time speed of 54 FPS.


\begin{table}[!t]\normalsize
    \caption{State-of-the-art comparison on the RGBD1K dataset. Our method (DTTrack) demonstrates superior generalization, matching the heavy teacher's F-score while achieving higher Precision and 5$\times$ faster inference speed.
    }
\label{tab:rgbd1k_sota}
\vspace{-2mm}
  \centering
\resizebox{1.0\linewidth}{!}{
  \setlength{\tabcolsep}{4mm}{
    \small
    \begin{tabular}{l|ccc|c}
    \toprule
    \multirow{2}*{\textbf{Method}} & \multicolumn{3}{c|}{\textbf{RGBD1K}} & \multirow{2}*{\textbf{FPS}}\\
        \cline{2-4}
 & RE & PR & F-score & \\
    \midrule[0.5pt]
    \multicolumn{5}{c}{\textit{State-of-the-Art Baselines}} \\
    \midrule
        SDSTrack & 48.5 & 50.0 & 49.3 & 28 \\
        UnTrack & 50.4 & 52.3 & 51.3 & 14 \\
        XTrack-B & 47.1 & 48.3 & 47.7 & 8 \\
        SUTrack-B224 (Teacher) & \textbf{\textcolor{red}{53.8}} & 56.0 & \textbf{\textcolor{red}{54.9}} & 11 \\
    \bottomrule
    \end{tabular}
    }
  }
 \vspace{-5mm}
\end{table}

\noindent \textbf{Attribute-based Performance Analysis.} To provide a granular view of our tracker's robustness, we analyze the performance across 12 distinct challenge attributes on the RGBT234 dataset, as visualized in Fig.~\ref{fig:radar}. These attributes include Fast Motion (FM), Partial/Heavy Occlusion (PO/HO), and Thermal Crossover (TC), among others. As observed in the radar chart, single-stream distillation methods (e.g., L2 and FNKD) exhibit noticeable performance fluctuations, falling short in challenging scenarios like Background Clutter (BC) and Scale Variation (SV). In contrast, our DTrack consistently envelops the performance curves of other strategies, achieving the outermost boundary across almost all axes. Notably, our method demonstrates a comprehensive advantage, maintaining high precision even in severe categories such as Deformation (DEF) and Low Illumination (LI). This holistic superiority confirms that coupling spatial structural alignment with semantic distribution alignment prevents the student model from overfitting to specific scenarios, thereby significantly enhancing robustness against diverse environmental variations.

\noindent \textbf{Component Analysis.} 
To isolate the contribution of each module, we conduct a component-wise ablation on the RGBT234 benchmark, as detailed in Table~\ref{tab:ablation_rgbt}. The baseline student, trained solely with ground-truth supervision, achieves an MPR of 86.9\%. 
First, applying generic feature-level constraints (L2) yields only a marginal improvement (+0.4\% MPR), suggesting that strict element-wise matching is overly rigid for transferring complex cross-modal representations. 
In contrast, our proposed components offer distinct advantages: employing Spatial Representation Alignment alone boosts the MPR to 89.2\%, verifying the necessity of transferring structural attention. Even more strikingly, utilizing Semantic Distribution Alignment individually propels the performance to 90.8\% MPR, highlighting the value of aligning decision boundaries. 
Crucially, the optimal performance is achieved only when these strategies are coupled. Our full framework reaches a peak MPR of 91.5\% and MSR of 67.0\%. This result empirically proves that spatial structural cues and semantic discriminative knowledge are not redundant but complementary; they function synergistically—spatial alignment sharpens the tracker's focus while semantic transfer refines the classification logic—enabling the lightweight student to fully approximate the teacher's superior capabilities.

\begin{table}[!t]\normalsize
    \caption{Ablation study of different distillation components on the RGBT234 dataset. Our strategy significantly outperforms individual components.
    }
\label{tab:ablation_rgbt}
  \centering
\resizebox{0.9\linewidth}{!}{
  \setlength{\tabcolsep}{5mm}{
    \small
    \begin{tabular}{l|cc}
    \toprule
    \multirow{2}*{\textbf{Distillation Strategy}} & \multicolumn{2}{c}{\textbf{RGBT234}} \\
        \cline{2-3}
 & MPR & MSR \\
    \midrule[0.5pt]
        No Distillation (Baseline) & 86.9 & 63.8 \\
        \midrule[0.1pt]
        + L2 (Generic) & 87.3 & 64.8 \\
        + (Spatial Only) & 89.2 & 66.3 \\
        + (Semantic Only) & 90.8 & 66.6 \\
        \midrule[0.5pt]
        \textbf{Ours} & \textbf{\textcolor{red}{91.5}} & \textbf{\textcolor{red}{67.0}} \\
    \bottomrule
    \end{tabular}
    }
  }
  \vspace{-3mm}
\end{table}

\section{Conclusion}
In this paper, we have presented DTrack, a unified multimodal tracking framework that effectively reconciles the longstanding conflict between high-performance fusion and real-time efficiency. 
Through a systematic analysis of architectural redundancy, we identified the prediction head as a critical yet often overlooked computational bottleneck. 
By strategically streamlining the decoder architecture via structural pruning, we unlocked real-time inference speeds of 54 FPS without compromising the encoder's rich feature representation. 
To bridge the resulting capacity gap between the lightweight student and the heavy teacher, we devised a novel Dual-Alignment Distillation framework. 
This approach synergizes Spatial Representation Alignment, which sharpens localization focus, with Semantic Distribution Alignment, which transfers discriminative decision boundaries, thereby ensuring comprehensive knowledge transfer. 
Our comprehensive evaluation across five benchmarks spanning RGB-Event, RGB-Thermal, and RGB-Depth modalities demonstrates that DTrack achieves state-of-the-art accuracy while delivering a $5\times$ speedup over the teacher model. 
Beyond immediate performance gains, this work establishes a new paradigm for efficient multimodal learning, proving that thoughtful architectural compression coupled with task-aware distillation can surpass heavy-weight baselines. 
We believe this provides a solid foundation for deploying robust multimodal trackers on resource-constrained edge devices, such as drones and autonomous robots. 


\bibliographystyle{IEEEtran}
\bibliography{main}

\end{document}